\documentclass[conference]{IEEEtran}
\IEEEoverridecommandlockouts
\usepackage{cite}
\usepackage{amsmath,amssymb,amsfonts}
\usepackage{algorithmic}
\usepackage{graphicx}
\usepackage{textcomp}
\usepackage{xcolor}
\usepackage[hidelinks]{hyperref}
\usepackage{hypcap}
\def\BibTeX{{\rm B\kern .05em{\sc i\kern .025em b}\kern .08em
    T\kern .1667em\lower.7ex\hbox{E}\kern .125emX}}
\begin{document}

\title{Defect-Aware Masked Autoencoders for Unsupervised Clustering of Steel Surface Defects}

\author{
\IEEEauthorblockN{ Shrey Patel}
\IEEEauthorblockA{
\textit{University of Maryland, College Park}\\
United States of America \\
ORCID: 0009-0003-0886-4583}}

\maketitle

\begin{abstract}
Reliable recognition of steel surface defects is essential for automated quality inspection, but supervised methods require manually annotated images that are costly to produce. This creates a need for methods that can learn useful defect representations directly from unlabelled surface images. This work presents Defect aware MAE, a label free masked autoencoder for clustering steel surface defects. Each image is divided into 196 patches, with 147 masked and 49 processed by a Vision Transformer encoder. Texture energy based masking preferentially hides structurally informative patches, while a shallow decoder reconstructs HOG descriptors that capture local texture and edge orientation. Strong augmentation and an intermediate layer readout are used to form the final image features. Pretraining uses no class labels, bounding boxes, detection branch, or contrastive loss. On the held out NEU DET split, UMAP followed by Ward clustering achieves 98.6\% Hungarian matched accuracy, 0.968 NMI, 0.967 ARI, and 0.838 silhouette score. Under an identical encoder and training schedule, it exceeds a SimCLR contrastive baseline in accuracy, NMI, ARI and silhouette scores across 5 seeds.
\end{abstract}


\section{Introduction}
Automated recognition of steel surface defects supports quality inspection, but supervised classification and detection methods depend on class labels or bounding box annotations~\cite{b3,b9,b11,b12}. Self supervised learning reduces this dependence by constructing training objectives directly from unlabelled images~\cite{b22}. Masked autoencoders learn visual representations by encoding visible image patches and reconstructing the missing information with a lightweight decoder~\cite{b4,b5}. Structured targets such as histograms of oriented gradients can direct masked models toward local texture and edge orientation rather than raw pixel intensity~\cite{b19,b20}.

This work presents Defect aware MAE, a label free masked autoencoder for clustering the six NEU DET defect categories~\cite{b1,b2}. The method combines HOG reconstruction, texture energy based masking, data augmentation, a shallow decoder and an intermediate layer feature readout. Pretraining uses no class labels, bounding boxes, detection branch or contrastive loss. Frozen features are projected with UMAP~\cite{b6} and grouped through Ward clustering. On 360 held out images never seen during pretraining, Defect aware MAE achieves 98.6 ± 0.2\% Hungarian matched accuracy across 5 seeds, while also producing higher NMI, ARI and silhouette values than a SimCLR baseline~\cite{b22} trained with the identical encoder and schedule.

\section{Literature Review}
Song and Yan introduced the NEU surface defect dataset with six categories of hot rolled steel defects and investigated texture based recognition through completed local binary patterns~\cite{b1}. Lv \textit{et al.} later provided a metallic surface defect benchmark and a deep detection network~\cite{b2}. Subsequent research has largely treated steel defect recognition as a supervised classification or detection problem. Examples include hierarchical feature fusion~\cite{b11}, Swin Transformer based classification and detection~\cite{b9}, broader deep learning classification methods~\cite{b12}, and MESC DETR, which combines a ConvNeXtV2 based backbone, edge enhanced feature fusion, and an adapted regression loss for metal defect detection~\cite{b3}. These methods depend on labelled classes or spatial annotations during training.

Self supervised learning offers a route for reducing this dependence on annotation. Xu \textit{et al.} combined self supervised representation learning with a segmentation network for surface defect detection~\cite{b10}. Hu \textit{et al.} applied contrastive representation learning with an Earth Mover's Distance matching metric to steel surface defects and transferred the pretrained encoder to supervised detectors~\cite{b25}. SimCLR provides a general contrastive framework in which two augmented views of each image are encoded and optimized through an agreement objective~\cite{b22}. Although these approaches reduce the amount of labelled data required for representation learning, their learned features are primarily evaluated through supervised detection or segmentation tasks~\cite{b10}.

Masked image modelling provides a noncontrastive alternative. The masked autoencoder encodes a small visible subset of image patches and reconstructs the missing content with an asymmetric lightweight decoder~\cite{b4}. Its encoder follows the Vision Transformer architecture, which represents an image as a sequence of patch tokens processed through self attention~\cite{b5}. Autoencoder based representation learning has also been applied beyond visual inspection, for example to predict volumetric deformation from InSAR interferograms~\cite{b18}. MaskFeat extends masked prediction by reconstructing structured visual features and reports that HOG targets provide an effective representation of local gradient structure~\cite{b19}. HOG describes local appearance through distributions of gradient magnitude and orientation~\cite{b20}. Stochastic Gumbel top k sampling further provides a method for selecting elements without replacement according to weighted probabilities~\cite{b21}. These ideas support the proposed combination of HOG reconstruction and texture energy guided masking.

Interpretation methods provide an additional way to examine learned visual representations. Grad CAM uses activation gradients to produce spatial importance maps~\cite{b7}, while Grad CAM++ refines the weighting of individual activations to improve localization~\cite{b8}. Both methods have also been applied to explain deep learning models in medical imaging~\cite{b17}. MESC DETR applies Grad CAM++ to examine attention in a supervised metal defect detector~\cite{b3}. In the present work, Grad CAM and Grad CAM++ are instead applied to a frozen label free encoder by targeting similarity between an image feature and its assigned cluster centroid. The resulting features are projected with UMAP~\cite{b6} and grouped without label supervision, while annotations are used only to evaluate clustering and explanation maps.

\section{Proposed Methodology}
The proposed Defect aware MAE uses a Vision Transformer based masked autoencoder to learn steel surface defect representations without class labels or spatial annotations. The released model was trained for 1,600 epochs with batch size 64. Training took 51 minutes on a single NVIDIA RTX A4000 GPU. The implementation, configurations, stored results, and reproduction instructions are publicly available at \url{https://github.com/ShreyPatel1311/NEU-Defect-Aware-Autoencoder}. The released model weights are available at \url{https://huggingface.co/Godseye1311/defect-aware-MAE}. The NEU DET dataset is publicly available on Kaggle.

\subsection{Data Collection and Augmentation}\label{sec:data}
\begin{figure}[htbp]
\centerline{\includegraphics[width=0.5\textwidth]{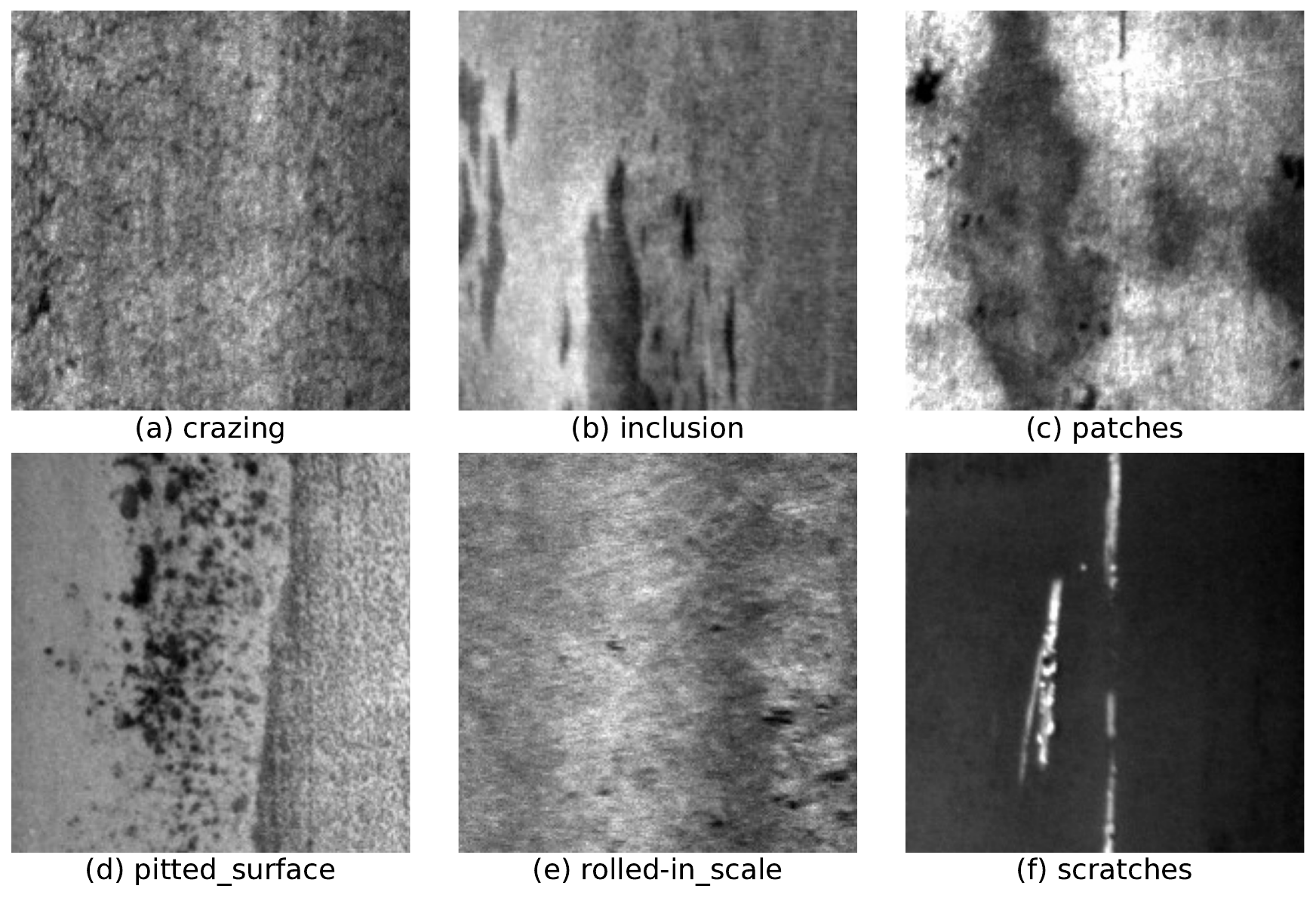}}
\caption{Sample of each labelled image present in the NEU-DET dataset}
\label{fig:dataset}
\end{figure}

The NEU DET dataset contains 1,800 grayscale images across six defect categories, with 300 images per category~\cite{b1,b2}. The distributed split contains 1,440 training images and 360 held out validation images. Pretraining uses only the training images and does not use labels or bounding boxes. The sample images of each defect are represented in Figure~\ref{fig:dataset}.

Each 200 $\times$ 200 image is enhanced using CLAHE~\cite{b23}, resized to 224 $\times$ 224 pixels, and scaled to the interval from 0 to 1. Training augmentation includes a random resized crop with an area fraction between 0.35 and 1, horizontal and vertical flips, rotations by multiples of 90 degrees, brightness and contrast adjustment, Gaussian blur, and Gaussian noise.

\subsection{Model Pretraining}

Defect aware MAE follows the asymmetric masked autoencoder framework~\cite{b4}. Each augmented grayscale image $x \in \mathbb{R}^{1 \times 224 \times 224}$ is divided into $N=196$ non overlapping patches of size $16 \times 16$. The method masks 147 patches and passes the remaining 49 patches to the encoder. A 2 layer decoder then predicts the HOG descriptor of every masked patch. Pretraining minimizes the reconstruction error at masked positions only.

\subsubsection{Encoder}

The encoder follows the Vision Transformer architecture~\cite{b5}. Each image patch $x_p^i$ is projected into a 768 dimensional token by a convolutional patch embedding. A learnable class token $x_{\mathrm{cls}}$ is prepended, and learnable positional embeddings $E_{\mathrm{pos}}$ are added:

\begin{equation}
z_0 =
[\,x_{\mathrm{cls}};\,
x_p^1 W_E;\,
x_p^2 W_E;\,
\ldots;\,
x_p^N W_E\,]
+ E_{\mathrm{pos}},
\end{equation}

where $W_E$ denotes the patch projection. Positional embeddings are added before patch selection so that each visible token retains its spatial position.

The visible sequence is processed by 6 Transformer layers. Each layer contains pre normalization multihead self attention with 12 heads, followed by a 3,072 dimensional feed forward network with GELU activation. For layer $l$, the operations are

\begin{equation}
z_l' =
z_{l-1}
+
\operatorname{MSA}
\left(
\operatorname{LN}(z_{l-1})
\right),
z_l =
z_l'
+
\operatorname{MLP}
\left(
\operatorname{LN}(z_l')
\right).
\end{equation}

\subsubsection{Defect Aware Masking}

Mask selection is based on the texture energy of each patch. For patch $i$, its energy $e_i$ is defined as the standard deviation of its pixel intensities. The energy values are standardized within each image:

\begin{equation}
\hat{e}_i =
\frac{e_i-\mu_e}
{\sigma_e+\epsilon},
\qquad
\epsilon = 10^{-6},
\end{equation}

where $\mu_e$ and $\sigma_e$ are the mean and standard deviation of the 196 patch energy values.

Each patch receives a score combining standardized texture energy with Gumbel noise:

\begin{equation}
s_i =
\beta \hat{e}_i + g_i,
\end{equation}

\begin{equation}
g_i =
-\log
\left(
-\log u_i
\right),
\qquad
u_i \sim \mathcal{U}(0,1).
\end{equation}

The 147 patches with the highest scores are masked, while the 49 patches with the lowest scores remain visible. The released model uses $\beta=1$. This gives patches with higher texture energy a greater probability of being masked while retaining stochastic selection through the Gumbel term~\cite{b21}. With mask ratio $r=0.75$, the number of visible patches is

\begin{equation}
N_{\mathrm{visible}}
=
N-\lfloor rN \rfloor
=
196-147
=
49.
\end{equation}

\begin{figure}[htbp]
\centerline{\includegraphics[width=0.5\textwidth]{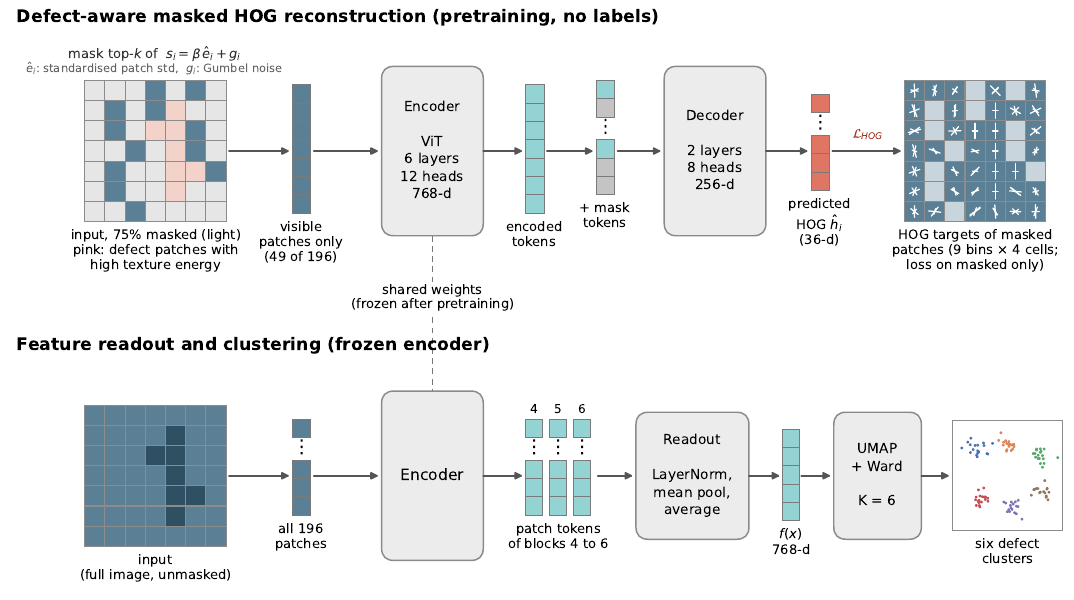}}
\caption{Model architecture of Defect aware MAE for predicting the HOG descriptors of masked patches}
\label{fig:Model}
\end{figure}

\subsubsection{HOG Reconstruction Target}

The reconstruction target is a histogram of oriented gradients rather than pixel intensity~\cite{b19,b20}. Horizontal and vertical Sobel filters produce gradients $G_x$ and $G_y$ for each pixel. Gradient magnitude and unsigned orientation are calculated as

\begin{equation}
m =
\sqrt{G_x^2+G_y^2},
\theta =
\operatorname{atan2}(G_y,G_x)
\bmod \pi.
\end{equation}

The orientation interval from $0$ to $\pi$ is divided into 9 bins. Each pixel contributes its gradient magnitude to the corresponding orientation bin. Histograms are averaged within $8 \times 8$ pixel cells. Since each $16 \times 16$ patch contains 4 cells, their histograms are concatenated into a 36 dimensional descriptor $h_i \in \mathbb{R}^{36}$, which is normalized as

\begin{equation}
h_i' =
\sqrt{36}\,
\frac{h_i}
{\lVert h_i \rVert_2+\epsilon}.
\end{equation}

\subsubsection{Decoder and Reconstruction Loss}

The decoder projects the encoded tokens from 768 to 256 dimensions. A shared learnable mask token is inserted at every masked position, and decoder positional embeddings are added to the complete sequence. The sequence is processed by 2 pre normalization Transformer layers with 8 attention heads. A final linear layer predicts a 36 dimensional HOG descriptor for each patch.

Let $\mathcal{M}$ denote the set of masked patch indices, $h_i'$ the HOG target, and $\hat{h}_i$ the decoder prediction. The training objective is the mean squared reconstruction error over masked positions:

\begin{equation}
\mathcal{L}_{\mathrm{HOG}}
=
\frac{1}{|\mathcal{M}|}
\sum_{i \in \mathcal{M}}
\frac{1}{36}
\left\|
\hat{h}_i-h_i'
\right\|_2^2.
\end{equation}

Only one augmented view of each image is required. Class labels, bounding boxes, detection losses, and contrastive losses are not used during pretraining. The model architecture is shown in Figure~\ref{fig:Model}.

\subsubsection{Optimization}

\begin{figure}[htbp]
\centerline{\includegraphics[width=0.5\textwidth]{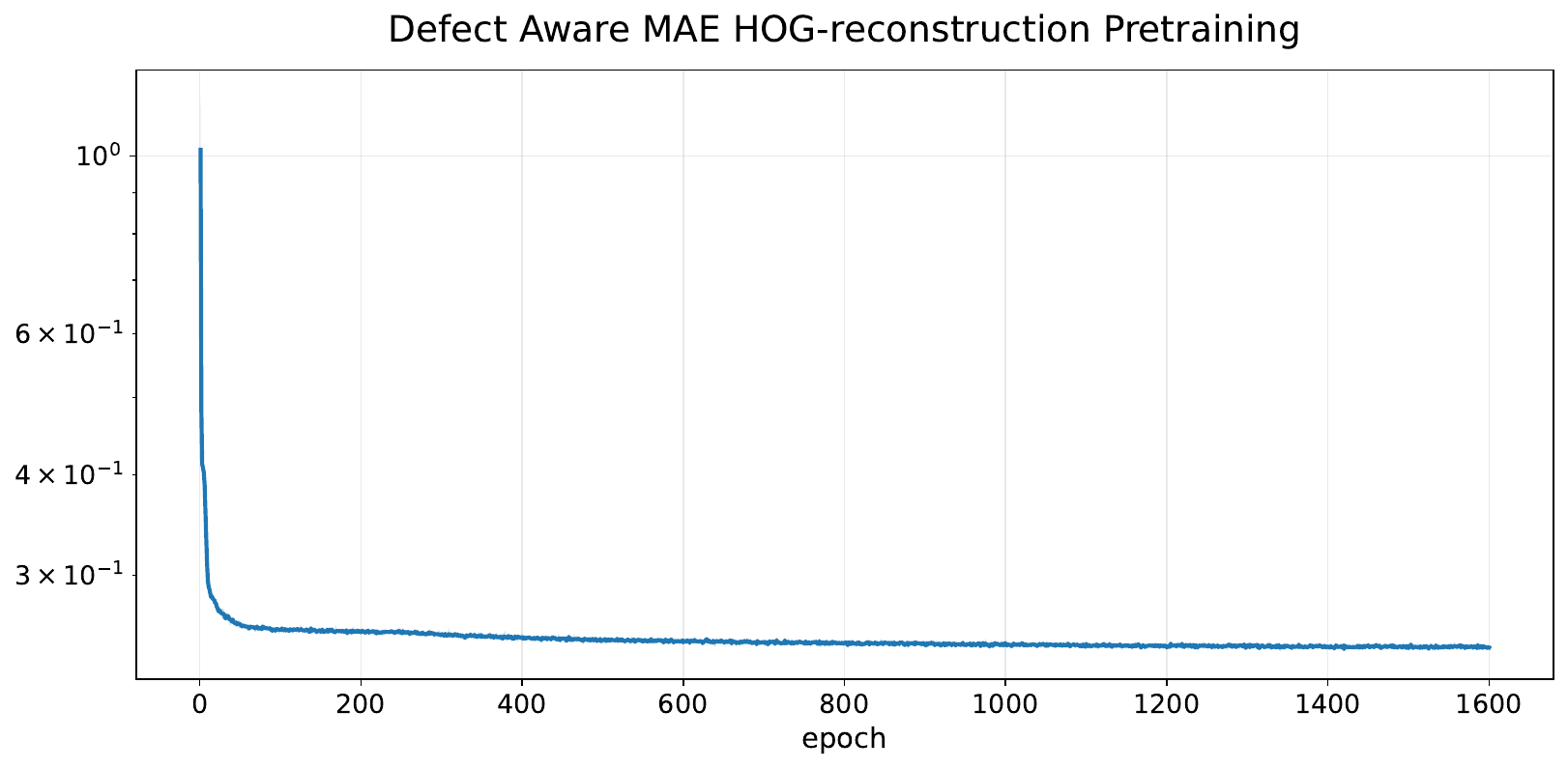}}
\caption{Masked HOG reconstruction loss and learning rate during pretraining}
\label{fig:training}
\end{figure}

Training minimizes the masked HOG reconstruction loss through one backward pass per batch. The model is trained for 1,600 epochs, with batch size 64 using AdamW~\cite{b24}, with $\beta_1=0.9$, $\beta_2=0.95$, a peak learning rate of $1.5 \times 10^{-4}$, and weight decay of 0.05. The learning rate increases linearly during the first 40 epochs and then follows cosine decay to zero. Training uses mixed precision and gradient norm clipping at 3.0. Progress is recorded through the reconstruction loss and learning rate, as shown in Figure~\ref{fig:training}.

\subsection{Dimensionality Reduction and Clustering}

After pretraining, the encoder is frozen and each image is processed without masking. Defect aware MAE uses the patch tokens from encoder layers 4 to 6 rather than the class token. The tokens from each layer are normalized, averaged across the 196 patch positions, and then averaged across the 3 layers:

\begin{equation}
f(x)=
\frac{1}{3}
\sum_{l=4}^{6}
\frac{1}{N}
\sum_{i=1}^{N}
\operatorname{LN}
\left(
z_i^{(l)}
\right),
\qquad
f(x)\in\mathbb{R}^{768},
\end{equation}

where $z_i^{(l)}$ is the representation of patch $i$ from encoder layer $l$, $N=196$, and $\operatorname{LN}$ denotes layer normalization without learned affine parameters. The resulting feature matrix is standardized to zero mean and unit variance along each feature dimension.

\subsubsection{Dimensionality Reduction}

The standardized features of the 360 held out images are projected into 2 dimensions using UMAP~\cite{b6}. The projection uses cosine distance, 25 neighbours, a minimum distance of 0.1, and random seed 742. UMAP is fitted to the complete held out feature matrix before clustering.

\subsubsection{Clustering}

The 2 dimensional UMAP representation is partitioned into $K=6$ clusters through agglomerative clustering with Ward linkage. The value of $K$ corresponds to the number of defect categories and is the only class information provided to the clustering procedure. Ward clustering begins with individual samples and repeatedly merges the pair of clusters that produces the smallest increase in within cluster variance.

\subsubsection{Evaluation}

Ground truth labels are used only after clustering to measure agreement between the discovered clusters and the six defect categories. Since cluster indices have no direct correspondence with class indices, the Hungarian algorithm determines the one to one mapping $\pi$ that maximizes agreement~\cite{b14}. With ground truth label $y_i$ and cluster assignment $c_i$, Hungarian matched accuracy is

\begin{equation}
\operatorname{ACC}
=
\frac{1}{n}
\sum_{i=1}^{n}
\mathbf{1}
\left[
y_i=\pi(c_i)
\right].
\end{equation}

The Adjusted Rand Index measures agreement between pairs of samples while correcting for chance~\cite{b15}. Define

\begin{equation}
A=\sum_{ij}\binom{n_{ij}}{2},\quad
B=\frac{1}{2}\left[
\sum_i\binom{a_i}{2}+
\sum_j\binom{b_j}{2}
\right],\quad
\end{equation}
\begin{equation}
E=\frac{
\sum_i\binom{a_i}{2}
\sum_j\binom{b_j}{2}
}{
\binom{n}{2}
}.
\end{equation}

The Adjusted Rand Index is then

\begin{equation}
\operatorname{ARI}=\frac{A-E}{B-E},
\end{equation}

where $n_{ij}$ denotes the number of samples shared by cluster $i$ and class $j$, while $a_i$ and $b_j$ are the corresponding row and column sums.

Normalized Mutual Information measures the shared information between cluster assignments $C$ and ground truth labels $Y$~\cite{b16}:

\begin{equation}
\operatorname{NMI}(C,Y)
=
\frac{
2I(C;Y)
}{
H(C)+H(Y)
},
\end{equation}

where $I(C;Y)$ is mutual information and $H(\cdot)$ denotes entropy.

Cluster compactness and separation are measured through the silhouette coefficient, which does not use ground truth labels~\cite{b26}. The mean silhouette is

\begin{equation}
S=
\frac{1}{n}
\sum_{i=1}^{n}
\frac{
\beta_i-\alpha_i
}{
\max(\alpha_i,\beta_i)
},
\end{equation}

where, for sample $i$, $\alpha_i$ is its mean distance to the other samples in its assigned cluster and $\beta_i$ is its mean distance to the samples of the nearest other cluster.

The main evaluation reports accuracy, ARI, NMI, and silhouette for UMAP followed by Ward clustering. The same evaluation settings are applied to Defect aware MAE and the SimCLR baseline.

\section{Results and Discussions}

Defect aware MAE and SimCLR were evaluated on the 360 held out NEU DET images using UMAP followed by Ward clustering. Table~\ref{tab:main_results} reports the mean and standard deviation across 5 seeds. Defect aware MAE achieved higher mean accuracy, NMI, ARI, and silhouette than SimCLR.

\begin{table}[t]
\centering
\caption{Clustering results on the held out NEU DET split. Values are mean $\pm$ standard deviation across 5 seeds.}
\label{tab:main_results}
\resizebox{\columnwidth}{!}{%
\begin{tabular}{lcccc}
\hline
\textbf{Model} &
\textbf{Accuracy (\%)} &
\textbf{NMI} &
\textbf{ARI} &
\textbf{Silhouette} \\
\hline
SimCLR &
$97.8 \pm 1.3$ &
$0.955 \pm 0.021$ &
$0.951 \pm 0.027$ &
$0.736 \pm 0.019$ \\
Defect aware MAE &
$\mathbf{98.6 \pm 0.2}$ &
$\mathbf{0.968 \pm 0.003}$ &
$\mathbf{0.967 \pm 0.004}$ &
$\mathbf{0.838 \pm 0.019}$ \\
\hline
\end{tabular}%
}
\end{table}

The silhouette score measured after PCA with 50 components was around $0.50$ for Defect aware MAE and $0.20$ for SimCLR. A linear probe trained on frozen features achieved $99.5$\% accuracy for Defect aware MAE and $98.8$\% for SimCLR. These measurements indicate that the Defect aware MAE features form more separated clusters and are more linearly separable under the same evaluation protocol.

Table~\ref{tab:ablation} isolates the contribution of each design choice under the same held out evaluation. In the component ablation, all four models share the same learning rate, decoder and training length, and differ only in the reconstruction target and the masking strategy. Replacing pixel reconstruction with HOG targets raises accuracy from 76.9\% to 92.8\% and silhouette from 0.530 to 0.802. Adding defect aware masking to the HOG target raises accuracy further to 96.4\%, with NMI and ARI increasing from 0.888 and 0.849 to 0.921 and 0.916. The two components are complementary: defect aware masking is most effective when the decoder predicts texture descriptors, as the masked high energy patches must then be reconstructed as edge and orientation structure rather than as pixel intensities.

The incremental configuration starts from a standard masked autoencoder with pixel targets and random masking, which reaches 73.9\% accuracy. HOG reconstruction with defect aware masking raises accuracy to 96.7\%. Strong augmentation increases the silhouette from 0.817 to 0.849, a 2 layer decoder raises accuracy to 97.9\%, and training for 1,600 epochs gives the final 98.6\% with NMI 0.968 and ARI 0.967.

\begin{table}[t]
\centering
\caption{Ablation study on the held out NEU DET split. Values with $\pm$ are mean $\pm$ standard deviation across seeds. All models use the intermediate layer readout.}
\label{tab:ablation}
\resizebox{\columnwidth}{!}{%
\begin{tabular}{lccccc}
\hline
\textbf{Configuration} &
\textbf{Seeds} &
\textbf{Accuracy (\%)} &
\textbf{NMI} &
\textbf{ARI} &
\textbf{Silhouette} \\
\hline
\multicolumn{6}{l}{\textit{Component ablation}} \\
\hline
Pixel target, random masking & 1 & 76.9 & 0.701 & 0.631 & 0.530 \\
Pixel target, defect aware masking & 1 & 71.9 & 0.747 & 0.638 & 0.624 \\
HOG target, random masking & 1 & 92.8 & 0.888 & 0.849 & 0.802 \\
HOG target, defect aware masking & 1 & \textbf{96.4} & \textbf{0.921} & \textbf{0.916} & \textbf{0.824} \\
\hline
\multicolumn{6}{l}{\textit{Incremental configuration}} \\
\hline
Standard MAE (pixel target, random masking) & 1 & 73.9 & 0.772 & 0.667 & 0.584 \\
+ HOG target and defect aware masking & 1 & 96.7 & 0.932 & 0.923 & 0.817 \\
+ strong augmentation & 5 & $96.9 \pm 1.8$ & 0.938 & 0.930 & 0.849 \\
+ 2 layer decoder & 3 & $97.9 \pm 0.3$ & 0.950 & 0.950 & \textbf{0.855} \\
+ 1,600 epochs (Defect aware MAE) & 5 & $\mathbf{98.6 \pm 0.2}$ & \textbf{0.968} & \textbf{0.967} & 0.838 \\
\hline
\end{tabular}%
}
\end{table}

\begin{figure}[htbp]
\centerline{\includegraphics[width=0.5\textwidth]{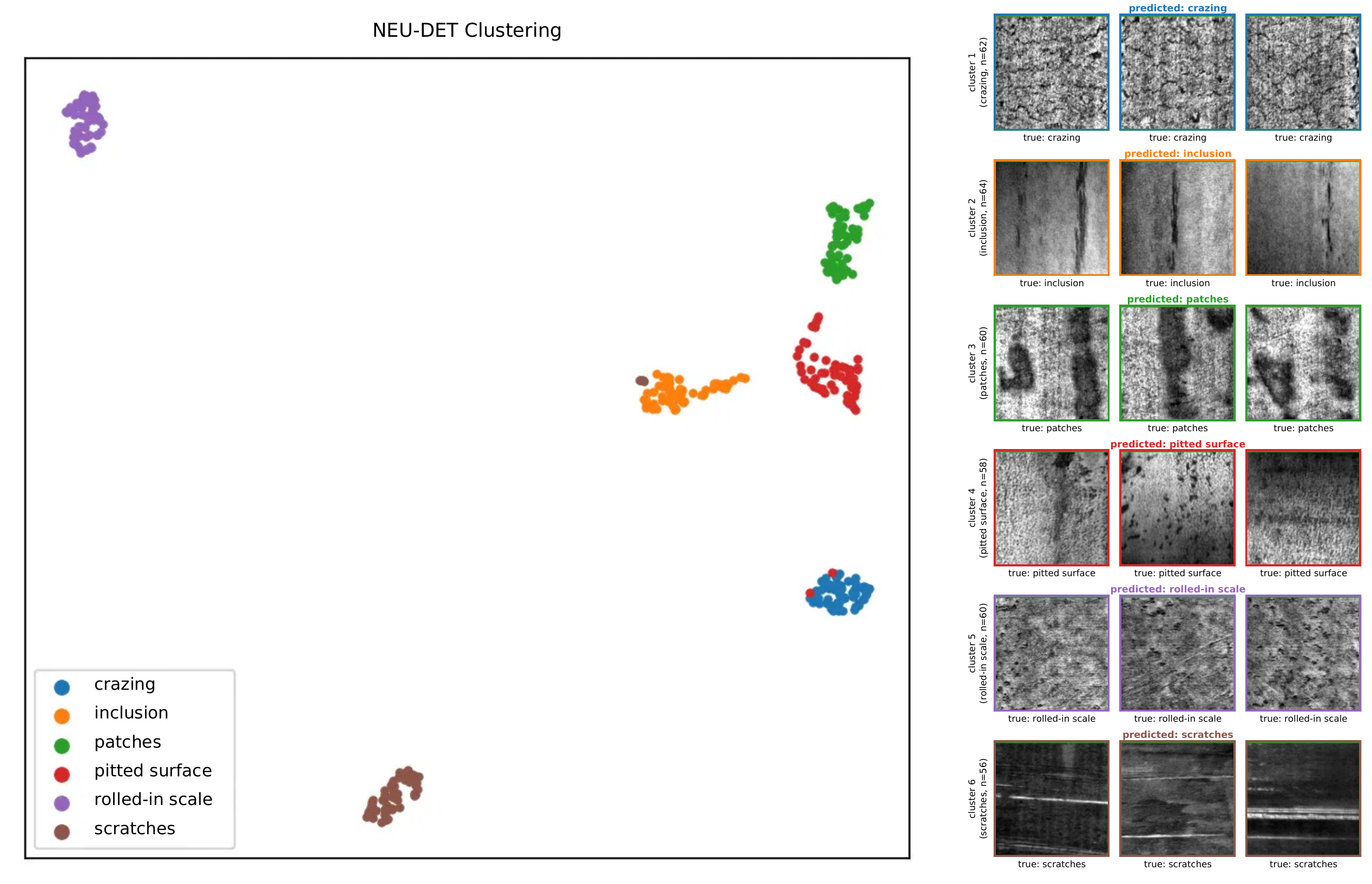}}
\caption{UMAP representation and Ward clusters of the released Defect aware MAE checkpoint}
\label{fig:clustering}
\end{figure}

Figure~\ref{fig:clustering} shows the UMAP representation of the trained encoder. The six Ward clusters correspond to the six defect categories after Hungarian matching. The model achieved 98.33\% accuracy, 0.963 NMI, 0.961 ARI, and a silhouette score of 0.827.

\begin{figure}[htbp]
\centerline{\includegraphics[width=0.4\textwidth]{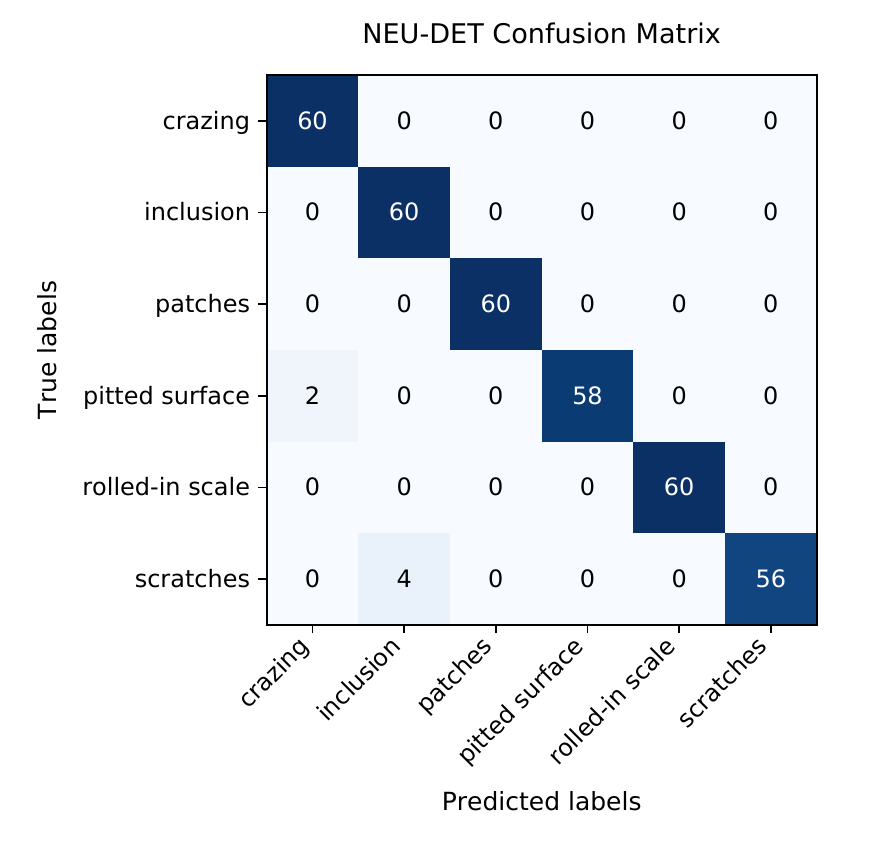}}
\caption{Confusion matrix between ground truth labels and predicted labels for the Defect aware MAE.}
\label{fig:conf-mat}
\end{figure}

The corresponding confusion matrix is shown in Figure~\ref{fig:conf-mat}. All 60 images were assigned correctly for crazing, inclusion, patches, and rolled in scale. For pitted surface, 58 images were assigned correctly and 2 were assigned to crazing incorrectly. Moreover, for scratches, 56 images were assigned correctly and 4 predictions were inaccurately clustered as inclusion. In total, 354 of the 360 held out images were assigned to their true labels precisely.

The encoder representation was examined with Grad CAM and Grad CAM++ using similarity between each image feature and its assigned cluster centroid as the explanation target. No classification head or class label was used to generate the maps. Across all 360 held out images, Grad CAM produced an energy in box value of 0.405, an energy in box lift of 1.15, and a pointing game score of 0.43. Grad CAM++ produced corresponding values of 0.394, 0.98, and 0.38. Deletion scores were 0.640 for Grad CAM and 0.642 for Grad CAM++, compared with 0.664 under random patch order.

\begin{figure}[htbp]
\centerline{\includegraphics[width=0.4\textwidth]{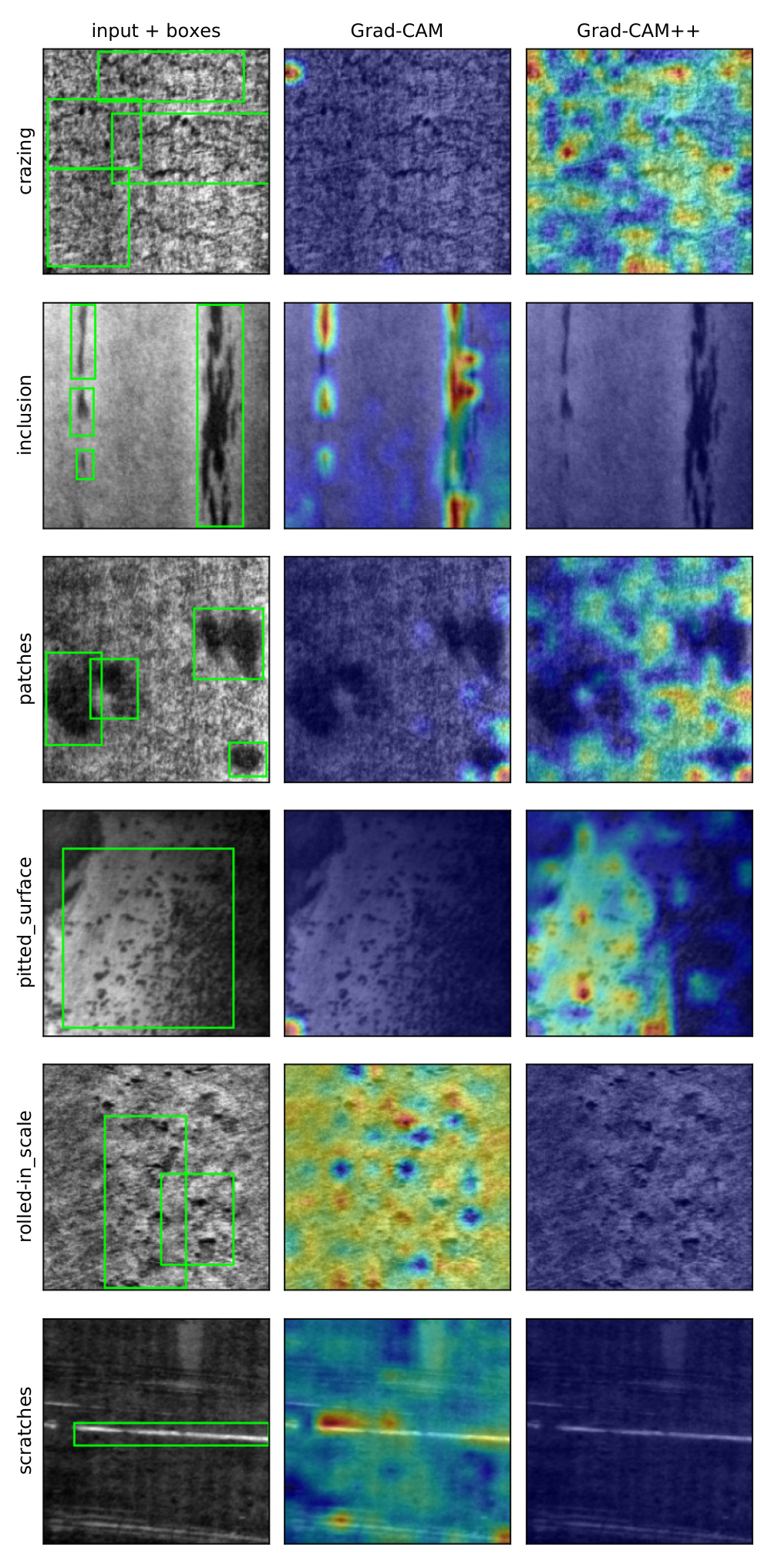}}
\caption{Input images with annotations and explanation maps produced by Grad CAM and Grad CAM++ for the Defect aware MAE}
\label{fig:grad-cam}
\end{figure}

Figure~\ref{fig:grad-cam} presents one example from each defect category. In the inclusion example, the Grad CAM response follows the vertical defect regions, while the scratches example places its main response along the horizontal scratch. Grad CAM++ distributes its response across the crazing texture and the pitted surface region. The patches and rolled in scale examples show responses distributed across multiple textured regions. Across each class, Grad CAM achieved an energy in box lift of 2.18 for inclusion, 1.49 for scratches, and 1.04 for rolled in scale. The pointing game score for pitted surface was 0.55 with Grad CAM and 0.57 with Grad CAM++, while the Grad CAM++ score for crazing was 0.58.

\section{Conclusion and Future Works}
This work presented Defect aware MAE, a label free masked autoencoder for clustering steel surface defects. The method combines HOG reconstruction, texture energy based masking, image augmentation, a 2 layer decoder, and an intermediate layer feature readout. Pretraining uses no class labels, bounding boxes, detection branch, or contrastive loss. Across 5 seeds, UMAP followed by Ward clustering achieved $98.6$\% Hungarian matched accuracy with $0.968$ NMI, $0.967$ ARI values and got $0.838$ silhouette score. Under the same encoder, data, optimizer, and learning rate schedule, SimCLR resulted in $97.8$\% accuracy with $0.955$ NMI, $0.951$ ARI, and $0.736$ silhouette score. Grad CAM and Grad CAM++ provided spatial maps of the image regions contributing to cluster similarity. These results show that masked HOG reconstruction with texture guided masking produces more accurate and stable clustering than the evaluated contrastive baseline.

\end{document}